\documentclass[runningheads]{llncs}
\usepackage{graphicx}
\usepackage{caption}
\begin{document}
\title{The Emotional Voices Database: Towards Controlling the Emotion Dimension in Voice Generation Systems}
\titlerunning{EmoV-DB: The Database of Emotional Voices}
%

\author{Adaeze Adigwe\inst{2}\thanks{These authors contributed equally to this work}
\and
No\'e Tits\inst{1}\textsuperscript{$\star$}
\and
Kevin El Haddad\inst{1}\orcidID{0000-0003-1465-6273} \and
Sarah Ostadabbas\inst{2} \and
Thierry Dutoit\inst{1}}
\authorrunning{A. Adigwe, N. Tits et al.}
%
\institute{num\'ediart Institute, University of Mons, 7000, Belgium \and
Augmented Cognition Laboratory, Northeastern University, Boston, USA
\\
\email{\{noe.tits, kevin.elhaddad, thierry.dutoit\}@umons.ac.be}\\
\email{ostadabbas@ece.neu.edu}}
\maketitle              
\begin{abstract}
In this paper, we present a database of emotional speech intended to be open-sourced and used for synthesis and generation purpose. It contains data for male and female actors in English and a male actor in French. The database covers 5 emotion classes so it could be suitable to build synthesis and voice transformation systems with the potential to control the emotional dimension in a continuous way. We show the data's efficiency by building a simple MLP system converting neutral to angry speech style and evaluate it via a CMOS perception test. Even though the system is a very simple one, the test show the efficiency of the data which is promising for future work.

\end{abstract}
\section{Introduction}
One of the major components of human-agent interaction systems is the speech synthesis module. The state-of-the-art speech synthesis systems such as wavenet and tacotron\cite{Oord16wavenet,Wang17tacotron,Shen17tacotron2} are giving impressive results. They can produce, intelligible, expressive, even human-like speech. But, they cannot yet be used to control the emotional dimensionality in speech which is a crucial parameter in order to obtain human-like controllable speech synthesis system.

Although still being relatively neglected by the affective computing community, the interest for emotional speech synthesis systems has been growing for the past two decades. After the improvement parametric systems brought to this field \cite{kawanami03emoVC,kev15speechSmiles}, deep learning-based systems were also employed for such a task.

One of the problems in the emotional speech synthesis research community is the lack of open-source data available and the difficulty to collect them. In fact, to the best of our knowledge, no open-source emotional speech database for synthesis purpose and suitable for deep learning systems is available. In this paper, we try to tackle this problem. We present an open-source multi-speaker (5 different speakers) and multilingual (in North American English and Belgian French) database of emotional speech. The database contains enough data of good enough quality to train deep learning-based systems for speech control and generation applications as we show later in this paper. So, in this paper, we propose an emotional speech dataset version of the CMU-Arctic speech database~\cite{cmuArctic04speechDB} which has been one of the reference open-source databases for speech synthesis since the early 2000.

In what follows we will present the background introducing this work in Section~\ref{background}. We will then talk about the motivation behind this work in Section~\ref{motivations} and detail the content of our database in Section~\ref{db}. In order to validate our database we will use it in a voice transformation experiment. Indeed the data will be used to transform neutral to emotional voice using deep learning-based systems and the obtained results will be evaluated using a Comparative Mean Opinion Score test in Section~\ref{validation}.

\section{Background}
\label{background}
\subsection{Emotion Representations}
Emotions can be represented in different ways. A first representation, is Ekman's six basic emotion model~\cite{Ekman99basicEmotions} which identify anger, disgust, fear, happiness, sadness and surprise as six basic emotions from which the other emotions may be derived.
Emotions can also be represented in a multidimensional continuous space like in the Russels circomplex model~\cite{russel05circumplexModel} (valence and arousal being the currently most famous dimensions used).
A more recent way of representing emotions is based on ranking which prefer a relative preference method to annotate emotions rather than labeling them with absolute values~\cite{yannakakis17ordinalemotions}.

\subsection{Related Work}
Several open-source databases can be found but to the best of our knowledge, none is really suitable for speech synthesis purpose. In this section we will explain why and mention some examples.
The RAVDESS database emotional data for 24 different actors \cite{ravdess18emoDB}. The actors were asked to read 2 different sentences in a spoken and sung way in North American English. The spoken style was recorded in 8 different emotional styles: neutral, calm, happy, sad, angry, fearful, disgust, surprise. Each utterance was expressed at 2 different intensities each (except for the neutral emotion) and 2 times thus giving a total of 1440 files. A perception test was then undertook to validate the database on the emotional categories, intensity and genuineness.

The CREMA-D database \cite{cremad14emoDB} is similar to the RAVDESS. For this database, 12 different sentences were recorded by 91 different actors, for the 6 basic emotions:happy, sad, anger, fear, disgust, and neutral. Only one of the 12 sentences was expressed in 3 different intensities, for the other 11, the intensity was not specified. The authors report 7442 files in total. This database was also validated through perception tests and helped validate the emotion category and intensity.

Also similar to the previous ones, the GEMEP database~\cite{GEMEP12emoDB} is a collection of 10 French-speaking actors, recorded uttering 15 different emotional expressions at three levels of intensity, in three different ways: improvised sentences, pseudo-speech, and nonverbal affect bursts. This database counts a total of 1260 audio files. It was also validated through perception tests.

The Berlin Emotional Speech Dataset \cite{berlinEmo05emoDB} contains the recording of 10 different utterances by 10 different actors in 7 different emotions (neutral, anger, fear, joy, sadness, disgust and boredom) in German, making it a total of 800 utterances (counting some second version of some of the sentences). This database was, like the previous ones, validated using perception experiments.

These database are all of good quality but are not suitable for current state of the art speech synthesis purpose because of the limited diversity of sentences recorded.

The IMPROV~\cite{improv-17-busso} and IEMOCAP~\cite{iemocap-08-busso} databases both contain a large amount of diverse sentences of emotional data.
IEMOCAP contains audio-visual recordings of 5 sessions of dyadic conversations between a male and a female subjects. In total it contains 10 speakers and 12 hours of data. IMPROV contains 6 sessions from 12 actors resulting in 9 hours of audiovisual data. Both databases were evaluated in terms of category of emotions~\cite{Ekman99basicEmotions} and emotional dimensions~\cite{russel05circumplexModel} by several subjects.
However they are not suitable for synthesis purpose either because although the data is well recorded and post-processed it contains overlapping speech due to the data recording setup (dyadic conversation) and some external noise.

The CMU Arctic Speech Database~\cite{cmuArctic04speechDB} and the SIWIS French Speech Synthesis Database~\cite{siwis17speechDB} are collections of read utterances of phonetically balanced sentences in English and French respectively. The CMU-Arctic database contains approximately 1150 sentences recorded from each of 4 different speakers while SIWIS contains a total of 9750 utterances from a single speaker. These are database suitable for speech synthesis purpose as there is a large amount of different sentences recorded from a single speaker in noiseless environment. However the sentences are neutral and do not express any emotions.

The AmuS database contain audio data dedicated to amused speech synthesis\cite{AmuS17emoDB}. We showed in previous work \cite{kev15speechsmilesynth1,kev15speechLaugh2,kev15speechLaugh} that this database was well suited for amused speech synthesis. But AmuS contains data only for amused speech and not other emotions.

\section{Motivations}
\label{motivations}
This database's primary purpose it to build models that could not only produce emotional speech but also control the emotional dimension in speech.
The techniques to allow this are either text-to-speech like systems where the system would map a given text sentence to a speech audio signal or voice transformation systems where a source voice would be converted to a specific target emotional voice.
Considering this, it is obvious that a lot of data is required. One of the primary difficulties of building emotional speech-based generation systems is the collection of data. Indeed not only must the recording be of good quality and noise free, but the task of expression emotional sentences in a large enough amount is challenging. Also it is often preferable concerning these types of systems, that a certain "class" of emotion contains data that are similar on the acoustic level.

The database presented here was built with these requirements in mind. The aim was also for it to fit with other currently open-source databases to maximize the quantity of data available.
As mentioned previously, the CMU-Arctic database (English) and the SIWIS (French) databases are two datasets of neutral speech. Each of them contain a relatively large amount of data that can be used as source voices for a voice conversion system or as pre-training data for a system. They are also transcribed which makes the transcription also available for our database. The transcribed utterances as well as annotations at phonetic level are available. A subset of these were used to build our database. The phonetic annotations are not time-aligned with our data yet, but methods can be used such as forced alignment systems \cite{brognaux12trainAndAlign}.

We chose five different emotions: amusement, anger, sleepiness, disgust and neutral. These emotions were chosen because of the ease to produce them by actors and in order to cover a diverse space in the Russel Circumplex.
This would help in later work to maybe generate other emotions not present in our database in Russel's model (using interpolation techniques for example \cite{kawanami03emoVC}).
 
\section{Database Content}
\label{db}
The data was recorded in 2 difference languages English (North American) and French (Belgian). English natives (2 females and 2 males) and a single male French native were asked to read sentences while expressing one of the above mentioned emotions. The English sentences were taken from the CMU-arctic database. The French ones from the SIWIS database. Both databases contain freely available open-source phonetically balanced sentences.

The recordings for the English data were carried on in 2 different anechoic chambers of the Northeastern University campus.
The ones for the French data were made in an anechoic room at the University of Mons.
All the data were originally recorded at 44.1k and were downsampled at 16k and stored in 16-bit PCM WAV format.

The utterances were recorded in several sessions of about 30 minutes recordings followed by a 5 to 15 minutes break and the data collection was spread across several days depending on the availability of the actors. The actors were asked to repeat sentences that were mispronounced.

The actors were asked to record each emotion class separately in different sessions. At the moment of redaction of this article, The sentences were segmented manually for some of the speakers (annotation and segmentation is still ongoing). By segmentation we mean determining the intervals of start and end of each sentence. The total number of utterances obtained is summarized in Table~\ref{table:utt}.

 \begin{table}
 \center
 \captionsetup{justification=centering}
 \caption{Gender and language of recorded sentences of/from each actor/speaker and amount of utterances segmented per speaker and per emotion. All speakers were recorded in all emotions, the - sign only signifies that the corresponding data were not segmented yet.}
 \label{table:utt}
 \begin{tabular}{|c|c|c|c|c|c|c|c|}
 \hline
 Speaker & Gender & Language & Neutral & Amused & Angry & Sleepy & Disgust\\
 \hline
 Spk-Je &  Female &English &417 &222 &523 &466 &189\\
 Spk-Bea &  Female &English &373 &309 &317 &520 &347\\
 Spk-Sa &  Male &English &493 &501 &468 &495 &497\\
 Spk-Jsh &  Male &English &302 &298 &- &263 &-\\
 Spk-No &  Male &French & 317 & - & 273 & - & -\\
 \hline
 \end{tabular}
 \end{table}

Amused speech can contain chuckling sounds which overlap and/or intermingle with speech called speech-laughs\cite{trouvain01speechlaughs} or can be only amused smiled speech~\cite{kev15speechSmiles}. So, for the amused data in our database, in order to collect as much data as possible and considering the relatively limited time the actors provided us, we focused on amused speech with speech-laughs. This choice was motivated by our previous study showing that this type of amused speech was perceived is perceived as more amused than amused smiled speech (without speech-laugh). Also in another study, we show that including laughter in synthesized is always perceived as amused no matter the style of speech it is inserted in (neutral or smiled) \cite{kev15speechLaugh}.
Based on the previous studies made on amusement, the actors were encouraged, while simulating the other emotions, to use nonverbal expressions before and even while uttering the sentences if they felt the need to (e.g. yawning for sleepiness, affect bursts for anger and disgust).

\section{Data Validation in a Voice Transformation Experiment}
\label{validation}
In order to validate our database, we show the performance of the data in a voice transformation system intended to generate target emotional speech from a source sentence. We thus designed an experiment described in this section.


\subsection{Voice Transformation System}
A feedforward-based voice transformation system was trained per speaker and per emotion. This experiment concerns Spk-Bea (female-English), Spk-Sa (male-English) and Spk-No (male-French). The system was trained to transform the neutral style (source) to another emotion style target. In this study, the target speech style is anger. This choice was motivated by the fact that the anger class was the one with the least nonverbal expressions which made the transformation task less complex for the simple system we chose. 
For each of the speakers a system was also trained to transform from neutral to neutral speech style. This was done to use the generated neutral utterances as reference and compare the emotional utterances generated to it. This thus gave us 6 systems trained in total.

The voice transformation system is based on the Merlin Toolkit~\cite{merlin-16-wu}. This toolkit contains a module allowing to perform voice conversion (VC), that is, transform a source speaker's voice so that it sounds like a target speaker's voice. The VC module do this by extracting speech features with a vocoder (the default WORLD vocoder~\cite{world-16-morise} was used) of both source and target voices, performing a Dynamic Time Warping (DTW) to align the features in time and computing a regression between the source and target features. The regression model used is a simple deep neural network (DNN) of 6 feedforward hidden layers in which each hidden layer is constituted of 1024 hyperbolic tangent units. 

In this experiment, instead of training the VC module with sentences uttered by a source and a target speaker, we trained it with sentences uttered by the same speaker with a source and a target emotion category. The procedure of this experiment is the same as in ~\cite{emo_vc-16-luo} which showed good results for emotion to neutral speech transformations.

Table~\ref{tab:trainingData} shows the amount of training data used to train each system. 

 \begin{table}
 \center
 \captionsetup{justification=centering}
 \caption{Amount of data used for training for neutral to neutral and neutral to angry tranformations}
 \label{tab:trainingData}
 \begin{tabular}{|c|c|c|c|}
 \hline
  Pairs & Spk-Bea & Spk-Sa & Spk-No \\\hline
 neutral-neutral &355 &456 &243\\\hline
 neutral-angry &296 &456 &243\\\hline
 \end{tabular}
 \end{table}


 \subsection{Perception Test}
 
 \begin{table}
 \center
 \captionsetup{justification=centering}
 \caption{Percentage of angry and neutral speech styles being accurately classified.}
 \label{tab:resultsEvalPerc}
 \begin{tabular}{|c|c|c|c|}
 \hline
 Pair & Spk-Bea & Spk-Sa & Spk-No \\
 \hline
 neutral-neutral & 96\% & 90\% & 98\%\\
 neutral-angry  & 78\% & 71\% & 83\%\\
 \hline
 \end{tabular}
 \end{table}

 \begin{table}
 \center
 \captionsetup{justification=centering}
 \caption{Mean and standard deviation of results obtained. Negative values would correspond to neutral being perceived as more emotional than the "anger" utterance, and vice cersa for the positive values. A "0" grade would indicate that there is no difference between the compared utterances.}
 \label{tab:resultsEvalRates}
 \begin{tabular}{|c|c|c|c|}
 \hline
 Pair & Spk-Bea & Spk-Sa & Spk-No \\
   & mean/std & mean/std & mean/std \\
 \hline
 neutral-neutral &0.05/0.2 &0.05/0.1 &0.01/0.09\\
 neutral-angry  &2.3/1.2 &2/2 &2.4/1.3\\
 \hline
 \end{tabular}
 \end{table}

After training, 5 neutral test sentences from the recorded data, not seen previously by the systems during training in each case, were transformed by each system to its target emotion. Then, each source-target pairs were used in a Comparative Mean Opinion Score~(CMOS) test. This makes it 30 different input-output pairs (including the neutral-neutral transformations).
Each pair was formed by the output of the system generating the neutral style with itself (neutral-neutral pair) or with the output of a system generating the angry style (neutral-angry pair) for the same speaker. Audio files were then created by concatenating both utterances to be compared in a single file with a 3 second silence delay between them. The order by which they were concatenated was random.
During this test, 26 participants per speaker were asked to grade on a scale of integers from -3 to 3 (0 included) which sentence was more emotional. They were then asked to pick among a list of 6 categories (neutral, sleepiness, anger, amusement, disgust or other) which one represented the most emotional utterance (0 corresponding to "no difference"). 
The more positive the value was, the more sure the participant considered that the first utterance in the audio clip was more emotional than the second utterance (and vice-versa for the negative). When processing the data, all ratings estimating that the neutral style was more emotional was converted to negative values and all others to positive (in case of the neutral-neutral pair all the ratings were therefore negative. Table~\ref{tab:resultsEvalPerc} shows the percentage of times the emotional utterances (utterance with highest grade or both if 0) were correctly categorized in each pair of speech styles.
The mean and standard deviation of the scores obtained by each pair are given in Table~\ref{tab:resultsEvalRates}.

We can see from the above tables that the participants could recognize the angry expression accurately and with high confidence for each speaker. These results show that the data for the "anger" emotion, can efficiently be used for a voice transformation system. 
It is interesting to note that most of the misclassification of the results in both the neutral-neutral case and the neutral-angry case were a due to perceiving the neutral (or one of the neutrals) as sleepiness or amusement. Indeed the test being a comparative one these misclassifications might be due to classifying one expression with respect to the other instead of in an absolute way. 

\section{Conclusion and Future Work}

In this paper we propose a first step towards obtaining a large open-source database of emotional data dedicated to systems aiming at controlling the emotional dimension in speech. We showed that the proposed database was efficient to produce angry voices from neutral ones using a simple DNN. In the near future we will continue collecting data and improve the database. Also, we will put this database into good use and build text-to-speech synthesis system using more complex systems than the ones used here, systems based on attention models or with architecture similar to wavenet. We hope that such systems will be efficient enough to learn not only the prosody representing the emotional voices but also the nonverbal expressions characterizing them which are also present in our database. \bibliographystyle{splncs04}
 \bibliography{mybib}
\end{document}